\pdfoutput=1

\documentclass[11pt]{article}

\usepackage[final]{acl}
\usepackage{times}
\usepackage{latexsym}

\usepackage[T1]{fontenc}

\usepackage[utf8]{inputenc}

\usepackage{microtype}
\usepackage{hyperref}

\usepackage{inconsolata}

\usepackage{graphicx}
\usepackage{caption}
\usepackage{subcaption}
\usepackage{xspace}

\usepackage[utf8]{inputenc}
\usepackage{booktabs}
\usepackage{microtype}
\usepackage{multirow}
\usepackage{rotating}
\newcommand{\datasetName}{{\sc TransientTables}\xspace}
\usepackage[nameinlink]{cleveref}
\usepackage[breakable]{tcolorbox}
\usepackage{tabularx}

\usepackage{graphics}
\title{
\datasetName: Evaluating LLMs' Reasoning on Temporally Evolving Semi-structured Tables} 

\author{
Abhilash Shankarampeta\textsuperscript{\rm 1}\thanks{~~Equal Contribution},
Harsh Mahajan\textsuperscript{\rm 2}\footnotemark[1] \\
\textbf{Tushar Kataria}\textsuperscript{\rm 2}, 
\textbf{Dan Roth}\textsuperscript{\rm 3},
\textbf{Vivek Gupta}\textsuperscript{\rm 4}\thanks{~~Corresponding Author}~\\ 
\textsuperscript{\rm 1}UC San Diego,
\textsuperscript{\rm 2}University of Utah, \\
\textsuperscript{\rm 3}University of Pennsylvania,
\textsuperscript{\rm 4}Arizona State University\\
ashankarampeta@ucsd.edu, harsh.mahajan@utah.edu, tkataria@cs.utah.edu\\ danroth@seas.upenn.edu, vgupt140@asu.edu \\
}

\begin{document}
\maketitle
\begin{abstract}
Humans continuously make new discoveries, and understanding temporal sequence of events leading to these breakthroughs is essential for advancing science and society. This ability to reason over time allows us to identify future steps and understand the effects of financial and political decisions on our lives.  However, large language models (LLMs) are typically trained on static datasets, limiting their ability to perform effective temporal reasoning. To assess the temporal reasoning capabilities of LLMs, we present the \datasetName dataset, which comprises 3,971 questions derived from over 14,000 tables, spanning 1,238 entities across multiple time periods. We introduce a template-based question-generation pipeline that harnesses LLMs to refine both templates and questions. Additionally, we establish baseline results using state-of-the-art LLMs to create a benchmark. We also introduce novel modeling strategies centered around task decomposition, enhancing LLM performance.

\end{abstract}

\section{Introduction}
\begin{figure}[ht]
\centering
\includegraphics[width=0.48\textwidth]{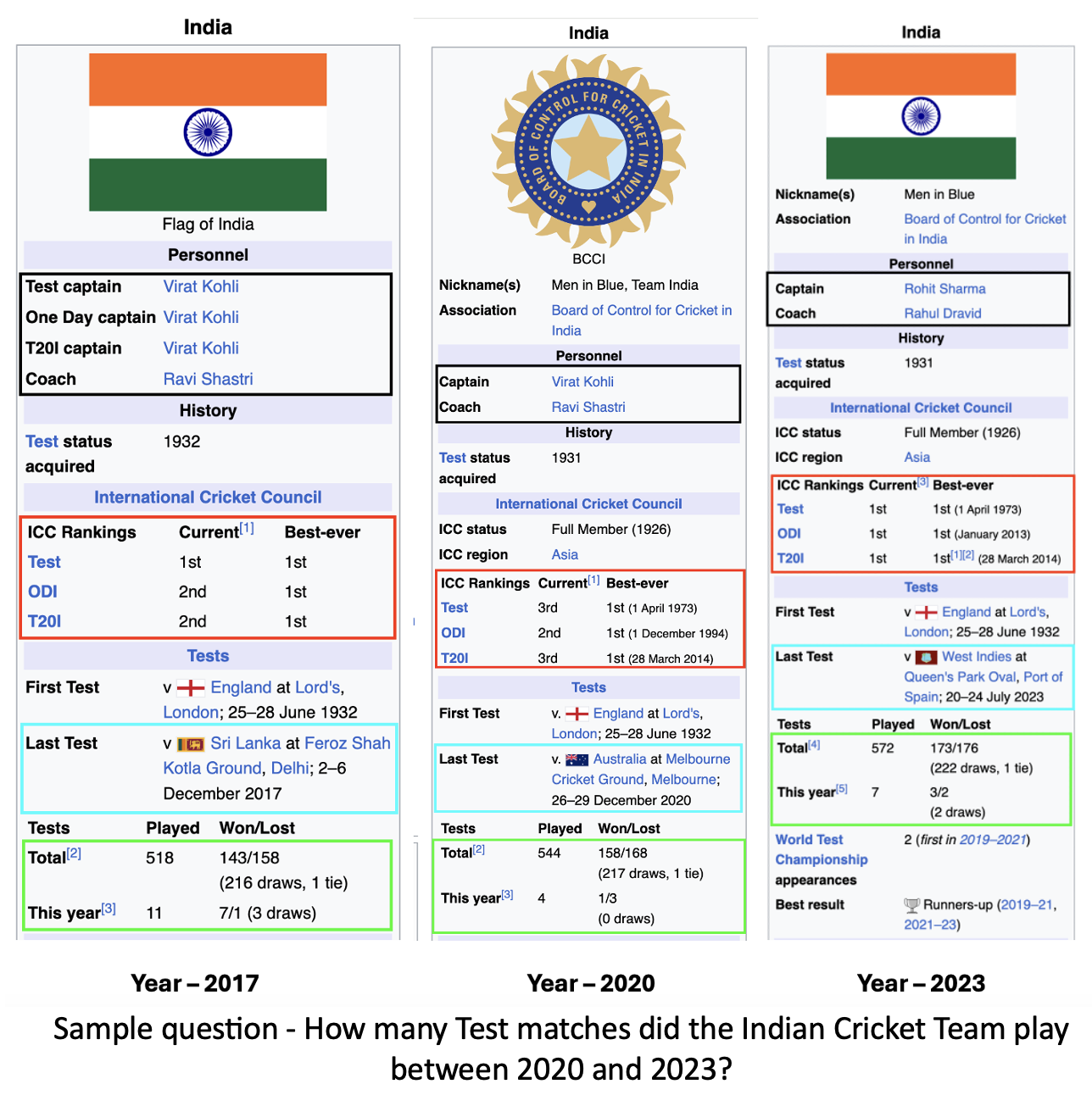}
\caption{\small \textbf{Example of Transient Information in Tables.} This example of the \textit{Indian Cricket Team} presents three tables sampled at different time points: 2017, 2020, and 2023. It clearly illustrates how certain values, such as \textit{Captain, ICC ranking, Tests played }, change over time. However, inconsistencies exist in the tables, including missing keys and incorrect values, such as the \textit{test status acquired} field, as noted in \citet{khincha-etal-2023-infosync}. In this work, we are only focusing on transient (or temporally changing) information.}
\label{fig:comparison-fig}
\vspace{-1.0em}
\end{figure}

In this day and age, information is constantly being updated depending on new facts and figures released in the public domain. 
Information is inherently transient and often subject to periodic or non-periodic updates. For instance, the profits, losses, and revenues of publicly traded companies fluctuate regularly, political figures shift with each election, bestseller lists change frequently, public transportation schedules are revised, rankings of women's football teams evolve, quarterly GDP growth varies, and the number of moons\footnote{\href{https://www.earth.com/news/its-official-earth-now-has-two-moons-captured-asteroid-2024-pt5/}{https://www.earth.com/news/its-official-earth-now-has-two-moons-captured-asteroid-2024-pt5/}} surrounding Earth can change based on new discoveries or pertinent information. \emph{This constant evolution underscores the dynamic nature of information across several fields.}

Timely updated information not only enhances the reader's knowledge but also shapes their perception. Furthermore, fluctuations in certain data, such as inflation, housing prices, and the cost of living, can significantly impact the lives of individuals. Most large language models (LLMs) \cite{achiam2023gpt,touvron2023llama,dubey2024llama} rely on static information, as they are trained on data that does not dynamically update. Given that retraining or finetuning these models is often costly, it is crucial to explore whether LLMs can effectively reason over temporal changes in information through in-context learning \cite{dong2022survey,gupta2023multi}. By incorporating temporal reasoning capabilities, LLMs could become more versatile, empowering them to handle a wider array of tasks.

Semi-structured tables are everywhere in this modern world, from web pages (see Figure \ref{fig:comparison-fig}) to nutrition labels on food products. Semi-structured tables such as Wikipedia Infoboxes (entity-centric) combine elements of both structured data (databases) and unstructured text. They offer structure while storing information in implicit forms, making them more flexible than traditional databases. By presenting information in a systematic, organized manner, tables allow for easy updates while maintaining organization and historical context, making them ideal for documenting the constant flux of dynamic information about entities such as public figures, corporations' revenues, and scientific concepts, among others \cite{gupta-etal-2020-infotabs,neeraja-etal-2021-infotabskg}. 

Temporal reasoning is particularly challenging in natural language processing (NLP) tasks, due to the constant updation of information required as stated above. 
To understand transient information in the natural language, the model must understand explicit and implicit temporal relations and track an entity's changing attributes. Recent works like TempTabQA \cite{gupta-etal-2023-temptabqa} and TIQ \cite{10.1145/3589335.3651895} are both focused on temporal question-answering on tabular data. However, TempTabQA only considers entity tables where each entity has a single table containing temporal information, such as an athlete winning different medals (gold, silver, or bronze) over multiple years. TIQ explores temporal QA with implicit time constraints from various sources, including knowledge bases, text, and Infoboxes (e.g., Who was the captain of the Indian Test Cricket Team \textit{before} Rohit Sharma?). Neither of these studies investigates understanding entity-centric tables that contain information that evolves, such as multiple tables of a certain entity across time (as highlighted in Figure \ref{fig:comparison-fig}). To the best of our knowledge, we are the first to study this problem. Specifically, we ask: \textit{ Can current language models understand and reason about the temporally evolving information (both periodic and nonperiodic) in semi-structured tables ?}. To address this problem, we define a question-answering task and create an associated dataset, \datasetName, where the answers require understanding and reasoning across at least two distinct tables sampled from different time periods. Even when a question can be answered using a single table, the model must still identify the correct table from a set of input tables to provide an accurate answer.

\datasetName consists of a QA dataset and a thorough analysis of the performance of state-of-the-art LLMs on entity-centric tables where values of keys in the table change over time.  
The resulting dataset comprises 3971 questions generated from more than 14k tables associated with 1238 entities (averaging 11.42 dynamic temporal tables per entity timeline). 
Our results indicate that SOTA LLMs struggle with reasoning on many straightforward questions, which humans can easily answer with sufficient context. This highlights that LLMs still have significant challenges in reasoning with transient information. Additionally, our experiments reveal that simple prompting often proves ineffective; therefore, breaking tasks down into smaller, more manageable components is necessary to achieve improved outcomes \cite{khot2022decomposed,ma2024task,wang2024hierarchical}. Analysis of our experimental results indicates significant deficiencies in LLMs' capacity for rightful evidence extraction and reasoning. Even when presented with the right evidence, these models demonstrate poor performance across various reasoning tasks compared to scenarios where the entire context is provided, which requires both rightful evidence extraction and reasoning processes. These findings suggest that LLMs rely on spurious correlations rather than robust logical inference when formulating responses. This work makes the following contributions\footnote{Website: \href{https://transienttables.github.io}{https://transienttables.github.io}}:
\begin{itemize}
    \item \textbf{\datasetName Benchmark:} A Question-Answering dataset on temporally evolving information in tabular data.
    \item \textbf{In-depth Analysis:} We conducted extensive experiments with LLMs to benchmark their performance and analyze their shortcomings. In addition, human evaluations were carried out, revealing a significant performance gap that LLMs must address.
\end{itemize}


\section{\datasetName Dataset}

\datasetName consists of infobox tables from various categories, including countries, cricket teams, economies, government agencies, and more. Each category features multiple entities, such as the USA, India, and Kenya, in the `countries' category, with 7 to 12 infoboxes per entity that capture temporal changes to form a timeline. Figure \ref{fig:comparison-fig} illustrates an example from the `cricket teams' category, showing a timeline of three infobox tables for the Indian cricket team. The categories were selected based on the infobox size (with more than 10 keys) and the degree of temporal variation. We manually chose infobox templates that met these criteria, prioritizing those with more substantial changes in key-value pairs over time. We chose ten categories: \textit{cyclists, equestrian players, economic data of a country, government agencies, cricket teams, field hockey players, golfers, table tennis players, countries, and cricketers}.

\paragraph{Entity Timeline Selection.}

We extract the infoboxes from the latest Wikipedia page and older versions of the same page. The extracted set of tables provides us with multiple attributes (ex. \textit{captain, ICC ranking} in Figure \ref{fig:comparison-fig}) of the same entity changing/evolving over time.
We start by extracting the current table from the latest Wikipedia page. Then, we go through the \textit{update history} to extract the important or pivotal moments for the entity of the current page. This process enables the extraction of periodic information, like quarterly profits and losses for a company, and non-periodic information, such as ranking for a sporting nation, which can change arbitrarily. 

\paragraph{Timeline Cleaning and Filtering.}
As a result of our extraction procedure and criteria for filtering entities, the dataset initially averaged 15 tables per entity. However, some entities had a higher number of tables, requiring pruning to meet our target range of 8-12 tables per entity. This range was chosen to accommodate the token limits of current state-of-the-art LLMs.
Pruning was performed using selection criteria based on the degree of variation between successive tables. We established category-specific thresholds that represent the minimum number of modified keys required between consecutive tables to warrant inclusion in the timeline. These thresholds were determined through two key factors: (1) the expected number of naturally changing attributes within each category, and (2) empirical testing to optimize context coverage while avoiding both over-selection (which would create redundant context) and under-selection (which would miss important changes). For the cricket team category, where we tracked nine dynamic attributes (captain, coach, rank, number of Tests played, Test record, number of ODIs, ODI record, number of T20Is, and T20I record), we set the threshold to 3 modified keys. This means that if a table differs from its predecessor in at least 3 of the tracked attributes, we include it in the timeline. This threshold effectively captured significant team developments while filtering out minor updates like grammatical rectification, shuffling keys, sorting, etc. This approach ensured a balanced representation of tables in the final dataset, allowing for focused analysis while controlling data volume. In addition, extensive data cleaning was performed to address noise (due to vandalic edits \footnote{\url{https://en.wikipedia.org/wiki/Vandalism_on_Wikipedia}}) and remove other irrelevant content from the tables.

\paragraph{Query-Answer Generation.} To evaluate the reasoning capabilities of LLMs when presented with transient information in tables, we created a question-answering dataset in which answers cannot be derived from a single table alone. Instead, observers must reference at least two tables from the timeline provided to arrive at a correct answer. Furthermore, even when a question can be answered using a single table, the language models must identify the appropriate table, i.e., retrieval, within the given set.

Question-answer pairs are generated through a semi-automated approach utilizing predefined templates. We manually crafted templates for each category and employed automated scripts to populate the details and enhance the quality of the questions.
Figure ~\ref{fig:comparison-fig} illustrates a sample of infobox data for the Indian Cricket Team between 2017 and 2023, with key details highlighted in colored boxes. This example raises several questions about time-varying information such as: \textit{`Who served as the coach of the Indian Cricket Team during Rohit Sharma's captaincy?' and `What was the Indian Cricket Team's winning percentage in Test matches in 2020?'}. From these generic questions, we created generalized templates that can be used for all the entities for the cricket team category. This allowed us to scale the question-answer pair set. Example templates for cricket team entities are shown below:

\begin{itemize}
\small{
    \item Name the person(s) who served as the \textit{<coach/test-coach/odi-coach/batting-coach/bowling-coach/fielding-coach>} when \textit{<captain/test-captain/odi- captain/t20i-captain:value1>} was the \textit{<captain/test-captain/odi-captain/t20-captain:key1>}?
    \item Does the Indian Cricket Team have the best win percentage in the \textit{<test/odi/t20i> format in <year:value1> or <year:value2>}?
    }
\end{itemize}

For each category, 10-15 templates were manually crafted, producing a diverse set of relevant question-answer pairs for the dataset.
Manually defined templates and generated questions were further refined using GPT-4o's to correct grammatical errors and resolve any ambiguities introduced by the template-based QA generation process. Check out the examples of QA generation templates for the cricket category in \cref{appendix:qa_template}.

\paragraph{\datasetName Statistics.} The semi-automated QA generation pipeline produced a total of 3,971 questions sampled from 14,133 tables, encompassing 1,238 entities of interest. On average, approximately 11.42 tables were extracted per entity.

To further analyze the types of temporal questions, we categorized them as either \textit{explicit} or \textit{implicit}. \textit{Explicit questions} specifically request time or date-related information, allowing the model to directly retrieve the relevant tables using the provided temporal references and reason over the data to generate an accurate answer. In contrast, \textit{implicit questions} lack direct temporal cues. To address these, the model must first establish temporal grounding by identifying the relevant tables (a.k.a right evidence) associated with the question. It then reasons over the extracted tables to arrive at the correct answer. Our dataset includes 2,985 implicit questions and 986 explicit questions.

We conducted an in-depth dataset analysis, concentrating on the types of reasoning required to solve the questions. These reasoning types were classified into nine distinct categories, as detailed in Table \ref{tab:reasoning-split}.

\begin{table}[!htb]
\vspace{-0.5em}
\small
    \centering
    \begin{tabular}{l|c} 
    \toprule 
        \bf Reasoning Types & \bf \# of QAs\\ 
        \midrule
        Extract the correct table from table timelines & 1,118 \\ 
        Calculate percentage & 157 \\ 
        Determine temporal difference & 676 \\ 
        Evaluate multiple differences $\&$ comparison & 350 \\ 
        Count unique values & 832 \\ 
        Determine the minimum value in a set & 227 \\ 
        Calculate ratio & 64 \\ 
        Determine the maximum value in a set & 314 \\ 
        Compare and contrast extracted values & 233 \\ 
        \bottomrule
    \end{tabular}
    
    \vspace{-0.5em}
\caption{\small\textbf{Reasoning Splits.} Dataset Split according to different reasoning operations required to answer query correctly.}
    \label{tab:reasoning-split}
    \vspace{-1.0em}
\end{table}

Furthermore, we evaluated the complexity of the reasoning by examining whether the questions involved analyzing a single key over time or multiple keys. The latter adds a greater level of complexity to the reasoning process. Specifically, 2,113 questions necessitated reasoning over a single key to arrive at the correct answer, while 1,858 questions required reasoning across multiple keys. See \cref{tab:qa-datasets} for a comparison of \datasetName and other temporal QA datasets.


\section{Modelling Techniques}

In order to respond to a query posed on a set of transient tables, a human evaluator must reason through the following steps:

 \begin{itemize} 
\setlength\itemsep{-0.25em} 

\item \textbf{Temporal Grounding.} Accurately identify and retrieve/extract the relevant set of tables necessary to answer the question. This can be regarded as retrieval over temporal information.
\item \textbf{Attribute Selection.} Effectively filter the relevant attributes, such as infobox table keys, from the retrieved tables. This step exemplifies information extraction on semi-structured information.
\item \textbf{Analytical Reasoning.} Analyze the information (values) within the appropriate keys to derive the correct answer. This involves several reasoning types: numerical reasoning for interpreting data, temporal reasoning for time-related concepts, lexical reasoning for word meanings, domain-specific reasoning for specialized knowledge, and common-sense reasoning for inferences.
\end{itemize}

These sequence of operations is highly interdependent, leading to compounded errors as we progress through each step. To test LLM capabilities and find areas where LLM needs improvement, we define a compressive set of modeling techniques using instruction sets (prompts) with different granularity of information (context in terms of a number of tables given as input), and intermediate task decomposition, i.e., \textit{Temporal Grounding}, \textit{Attribute Selection} and \textit{Analytical Reasoning}.

\paragraph{Information Granularity Variations.} To evaluate whether LLMs can effectively ground their responses, we vary the granularity of contextual information provided to the model. We assess their reliance on pre-trained knowledge (static information acquired during training, with no contextual information) versus their ability to adapt to new information included in the query. By varying the granularity of contextual information, we further evaluate the model's reasoning ability. To achieve this, we define two distinct types of instruction sets:

\paragraph{- Closed Book.}  In this prompt, the language model is presented only with the question and must generate an answer without any additional context information (\textit{without tables}). The model relies entirely on its internal, pre-existing (parametric) knowledge to respond, functioning in a closed-book setting. In this scenario, the LLM must accurately recall its pre-trained knowledge to answer the question. 
    
\paragraph{- Open Book.} In this prompt, the language model is provided with various sets of tables as context to answer the questions, operating in an open-book setting. Here, the LLM needs to ground its responses in the provided information and reason effectively across multiple table timelines to answer the queries accurately. To assess the capabilities of the LLM, we further categorize the granularity of information into two scenarios:

\begin{itemize}
\setlength\itemsep{0.25em} 
    \item \textit{Single Table.} A randomly or latest (most recent in the timeline, i.e., the last entry) selected table from the extracted set is provided as an input prompt. This approach simplifies the task while limiting historical context. Here, we evaluate whether LLMs can reason about temporal information from a static data sample. 
    
    \item \textit{Full Timeline.} The complete timeline, comprising all tables extracted for the entity, is provided as input. . This seeting also test model ability to filter relevant information from broader contexts. The model must perform all three steps (temporal grounding, attribute selection, and analytical selection) to arrive at the correct answer.

    \item \textit{Oracle Timeline}: Only the most relevant tables (1-4) are provided, simulating perfect extraction to isolate reasoning from retrieval challenges. Here, the model must perform the last two steps i.e. attribute selection and analytical reasoing to arrive at the correct answer.
\end{itemize}

\paragraph{Task Decomposition.} Initial experiments using a straightforward instruction set to explain the task and various contextual variations revealed that LLMs struggle with accurate reasoning, resulting in poor performance. To improve LLM effectiveness, we developed prompts that \emph{pragmatically} break down the transient reasoning task into smaller, more manageable components, as previously outlined (i.e., temporal grounding, attribute selection, and analytical reasoning). To assess the effectiveness of different task decomposition strategies, we propose the following variations:

    \paragraph{- Without Decomposition:} This method employs a basic prompt that presents the task description alongside relevant in-context information, as previously outlined and instructs the model to generate an answer.
  
    \paragraph{- Intermediate Breakdown:} This method assesses LLMs in two key areas: (a) their ability to retrieve relevant tables essential for answering questions and (b) their proficiency in reasoning with those tables. This approach includes three variations:

\textit{(a.) Information Retrieval:} This two-stage QA approach comprises two steps: (1) \textbf{Table Retrieval}, in which the language model extracts relevant tables from the timeline necessary to answer the question, and (2) \textbf{Answer Generation}, where the model utilizes these extracted tables for reasoning.

\textit{(b.) Information Extraction:} This approach is a variant of table retrieval; however, instead of retrieving relevant tables, the model focuses on directly \textbf{extracting specific attributes, such as infobox keys, from tables relevant} to the query. The main distinction between the two methods lies in the granularity of the data structure being retrieved—tables versus individual extracted keys.

\textit{(c.) Information Retrieval-Extraction:} This three-stage method incorporates an additional step for a more granular approach: (1) \textbf{Table Retrieval}, in which the language model identifies and retrieves the relevant tables needed to answer the question; (2) \textbf{Attribute Extraction}, where the model extracts pertinent attributes, such as infobox table keys, from the extracted tables; and (3) \textbf{Answer Generation}, in which the model utilize the extracted keys to reason and derive the correct answer to the question.

These multi-stage approaches enable a more comprehensive evaluation of the LLM's capabilities at each step of the process. The evaluation is conducted across all variations of the context settings, i.e., without table, single table, full-timeline and oracle.

\paragraph{How to extract evidence?} Our setup includes 8–12 temporally ordered tables per entity, each forming a timeline of evolving attributes. Retrieving relevant tables and extracting key-attribute pairs is challenging due to the high semantic similarity among tables of the same entity. Capturing subtle temporal changes adds further complexity. Traditional methods like BM25 and dense retrievers, designed for diverse document collections, often struggle with fine-grained temporal distinctions. To overcome this, we leverage LLMs with tailored prompts for more effective retrieval and extraction.

\paragraph{Models Utilized:} For our evaluations, we employed the following models: Llama3-70B \cite{llama3modelcard}, GPT-4o, GPT-4o-mini, Gemini-1.5-flash \cite{reid2024gemini}, Llama3-8B, and Mixtral-8x7B \cite{jiang2024mixtral}. In the single-stage setting (without task decomposition), we applied various prompting techniques, including Zero-shot, Few-shot, and Few-shot with Chain-of-Thought. For the multi-stage setting (with task decomposition prompts), we utilized Zero-shot and Chain-of-Thought prompting methods. In our implementation, we converted all tabular data into JSON string format before passing them to the LLMs. Check out the prompts used in the experiments in \cref{appendix:prompts}.

\paragraph{Evaluation Metrics:} We employed several metrics to compare the results across different models: \textit{F1 score}, \textit{Exact Match (EM)}, \textit{Rouge-1 (R1)}, and \textit{Rouge-L (RL)}. The \textit{F1 score} and \textit{Exact Match (EM)} are reported in the main paper, while the other metrics are detailed in the appendix. These metrics are widely used for evaluating QA task performance.

\section{Results and Analysis}
Our experiments answer the following questions: 
\vspace{-0.5em}
\begin{itemize}
\setlength\itemsep{-0.35em}
    \item Is question answering over transient information a challenging task for current LLMs?
    \item Do closed-source API access models outperform open-source models, and to what extent?
    \item What impact does task decomposition have on performance improvement?
    \item Does fine-tuning the model on a subset of the dataset enhance its performance? If so, to what degree?
\end{itemize}
 
 \textbf{\datasetName is challenging.}
Tables \ref{tab:zero-shot}, \ref{tab:few-shot}, and \ref{tab:cot} demonstrate that reasoning with temporally evolving information poses significant challenges. The GPT-4o model achieves the highest F1-score and exact match scores of 63 and 58, respectively, when leveraging all tables as context and Chain of Thought (COT) prompting.  Humans achieve an F1-score of 93 and an exact match of 88, significantly outperforming the best models, with the top-performing model lagging by 30 F1 points. These results indicate that current state-of-the-art models struggle to effectively comprehend temporally evolving information. See Appendix \ref{subsec:human_eval_info} for the complete human evaluation procedure.

\begin{table*}[!htb]
     \centering
     \scalebox{0.64}{
     \begin{tabular}{p{2.3cm}@{}|c|rr|rr|rr|rr|rr|rr}
& &\multicolumn{2}{c|}{\bf GPT-4o} &\multicolumn{2}{c|}{\bf Llama3-70b} &\multicolumn{2}{c|}{\bf Gemini-1.5} &\multicolumn{2}{c|}{\bf GPT-4o-mini} &\multicolumn{2}{c|}{\bf Llama3-8b} &\multicolumn{2}{c
}{\bf Mixtral} \\\cmidrule{3-14}
\bf Context &\bf Decomposition &F1 &EM &F1 &EM &F1 &EM &F1 &EM &F1 &EM &F1 &EM \\\midrule
Without Table &- &  19.43& 14.92&11.9&7.13&13.74&9.15&14.54&10.27&9.75&6.28&9.43&5.12\\\midrule
Single Table&- &31.9& 28.4&26.59&22.21&26.06&23.16&30.07&26.3&25.41&21.27&12.38&7\\
Latest Table&- &35.22 & 31.42 & 26.59 & 22.21 & 28.43 & 25.5 & 32.79 & 29.16 & 25.41 & 21.27 & 18.7 & 14.96\\\midrule
\multirow{4}{*}{Full Timeline}&Without Decomposition(WD) & 46.12& 40.54&35.34&23.85&36.61&29.51&40.59&32.76&30.94&21.46&28.15&20.3\\
&Information Retrieval(IR)& 51.93&45.8& 41.14& 32.89&34.58&25.64&44.5&37.12& 19.08& 11.83& 25.75& 19\\
& Information Extraction(IE)&\bf 53.37&\bf 47.4&\bf 45.32&\bf 37.2&\bf 47.08&\bf 37.7& 46.51&39.3&\bf 35.39&\bf 28& \bf 29.45& \bf 22.19\\
 & Information Retrieval-Extraction(IRE)& 52.96&47.27& 45.08& 37.12& 42.92& 33.98&
\bf 47.83& \bf 41.18& 24.48& 17.68&  25.77&17.73\\\midrule
\multirow{2}{*}{Oracle Tables}&Without Decomposition & 56.88&53.56& 34.66& 24.78& 41.99& 35.56& 44.92&39.22& 32.59& 24.67& 23.26& 17.78\\

&Information Extraction &55.3& 51.67& 39.53& 35.67& 39.17& 35.56&
45.16&39.92& 11.61& 6.11& 24.38& 20.33\\
\end{tabular}}
\vspace{-0.5em}
\caption{\small\textbf{Zero Shot Results.} Results in different in-context variations and different intermediate task decompositions with zero-shot prompting. The number of tables as input is limited by token length, which limits the Full timeline to 12 tables.}
\vspace{-0.5em}
\label{tab:zero-shot}
\end{table*}

\begin{table*}[!htb]
     \centering
     \scalebox{0.67}{
     \begin{tabular}{p{2.3cm}@{}|c|rr|rr|rr|rr|rr|rr}
& &\multicolumn{2}{c|}{\bf GPT-4o} &\multicolumn{2}{c|}{\bf Llama3-70b} &\multicolumn{2}{c|}{\bf Gemini-1.5} &\multicolumn{2}{c|}{\bf GPT-4o-mini} &\multicolumn{2}{c|}{\bf Llama3-8b} &\multicolumn{2}{c
}{\bf Mixtral} \\\cmidrule{3-14}
\bf Context &\bf Decomposition &F1 &EM &F1 &EM &F1 &EM &F1 &EM &F1 &EM &F1 &EM \\\midrule
Single Table&- & 38.28& 34.27&27.86&24.13&34.05&30.04&35.31&30.92&25.85&22.35& 20.52& 20.48\\
Latest Table&- & 38.28 & 39.16 & 27.86 & 30.37 & 34.05 & 34.96 & 35.31 & 36.08 & 25.85 & 28.15 & 23.51 & 24.22\\\midrule
\multirow{4}{*}{Full Timeline}&Without Decomposition & 54.33& 49.26&42.26&33.62&47.52&40.3&43.73&37.01&28.22&20.08& 32.31 & 33.93\\
&Information Retrieval& 51.57& 46.5& 46.92& 39.2&\bf 47.79&\bf 40&\bf 45.24&\bf 38.7& 22.85& 15.3& 26.19& 27.36\\
& Information Extraction& 53.64& 48.2&47.83&39.4&46.45&36.9&43.17&34.4&\bf 34.11&\bf 26.3&\bf32.15&\bf33.47\\
 & Information Retrieval-Extraction& \bf 55.89&   \bf 50.6& \bf 48.04& \bf 40.6&  46.4&  37.5&  44.72&  36.4&  20.99&  22.78&  24.12&25.64\\\midrule
\multirow{2}{*}{Oracle Tables}&Without Decomposition &62.52& 59.22& 47.11& 42.11& 48.92& 43.89&49.58&44.56& 35.88& 31.33& 30.06& 24.22\\

&Information Extraction & 57.29& 53.78& 44.8& 40.78& 46.97& 42&47.01&41.44& 16.46& 11.33& 28.72& 23.89\\
\end{tabular}}
\vspace{-0.5em}
\caption{\small\textbf{Few Shot Results.} F1 and Exact Match scores for different in-context variations and intermediate task decompositions with few-shot prompting. The number of tables as input is limited by token length, which limits the Full timeline to 10 tables.}
\label{tab:few-shot}
\vspace{-0.5em}
\end{table*}

\begin{table*}[!htb]
     \centering
     \scalebox{0.65}{
     \begin{tabular}{p{2.3cm}@{}|c|rr|rr|rr|rr|rr|rr}
& &\multicolumn{2}{c|}{\bf GPT-4o} &\multicolumn{2}{c|}{\bf Llama3-70b} &\multicolumn{2}{c|}{\bf Gemini-1.5} &\multicolumn{2}{c|}{\bf GPT-4o-mini} &\multicolumn{2}{c|}{\bf Llama3-8b} &\multicolumn{2}{c
}{\bf Mixtral} \\\cmidrule{3-14}
\bf Context &\bf Decomposition &F1 &EM &F1 &EM &F1 &EM &F1 &EM &F1 &EM &F1 &EM \\\midrule
\multirow{4}{*}{Full Timeline}&Without Decomposition & 57.77& 51.92&51.9&44.54&52.91&44.59&49.06&41.79&\bf 39.54&\bf 31.55& \bf 33.98 & \bf 36.28\\
&Information Retrieval& 59.36& 54.8& 50.48& 43.5&44.61&34.1&48.57&41.3& 24.41& 16.8& 27.44& 28.65\\
& Information Extraction& 62.04& 57.5&\bf 55.46&\bf 47.6&53.74&46.1&42.81&31.8& 35.29& 27.5&32.17&33.19\\
 & Information Retrieval-Extraction&  \bf 65.06&  \bf 60.11&  53.94&  45.56& \bf 54.08& \bf 43& \bf 56.4& \bf 48.33&  22.93&  15.56&  28.56& 30.22\\\midrule
\multirow{2}{*}{Oracle Tables}&Without Decomposition & 59.29& 53.67& 50.61& 45.22& 51.63& 46.56&49.89&41.89& 30.72& 23.78& 26.16& 21.67\\

&Information Extraction & 60.44& 56.56& 48.67& 44& 48.23& 42.89&44.38&36& 16.27& 11.33& 15.28& 11.89\\
\end{tabular}}
\caption{\small\textbf{COT Results.} F1 and Exact Match scores for different in-context variations and intermediate task decompositions with COT prompting. The number of tables as input is limited by token length, which limits the Full timeline to 7 tables.}
\label{tab:cot}
\vspace{-1.5em}
\end{table*}

\begin{table}[!htb]
     \centering
     \scalebox{0.56}{
     \begin{tabular}{p{2.5cm}@{}|c|ll|rr|rr}
& Samples used  & \multicolumn{2}{c|}{\shortstack{\bf Without \\ \bf Fine Tuning}}&\multicolumn{2}{c|}{\bf 100} &\multicolumn{2}{c}{\bf 1000}     \\
\hline
\bf Context &\bf Decom.   & F1&EM&F1 &EM &F1 &EM \\\midrule
Without Tables&-   & 14.54&10.27& 17.94&13.02&\textbf{21.95}&\textbf{17.8}\\\midrule
Single Table&-   & 30.07&26.3&35.55&31.88&\textbf{42.41}&\textbf{39.3}\\
Recent Table&-  & 35.31&36.08&54.18&54.20&\textbf{75.84}&\textbf{75.4}\\\midrule
\multirow{4}{*}{Full Timeline}&WD   & 40.59&32.76&48.98&44.23&\textbf{67.06}&\textbf{63.2}\\
&IE  & 46.51&39.3&51.35 &46.9 &\textbf{73.95} &\textbf{70.3} \\
& IR  & 44.5&37.12&51.02 &46.8 &\textbf{74.64} &\textbf{71.2} \\
 & IRE  & 47.83&41.18&51.06 &46.5 &\textbf{73.8} &\textbf{70.1} \\
\end{tabular}}
\vspace{-0.25em}
\caption{\small\textbf{Zero-Shot Results with Fine Tuned GPT-4o-mini.} Results of various in-context and task decomposition settings with zero-shot prompting using fine-tuned models trained on 100 and 1000 samples.}
\label{tab:zero-shot-fine-tuned}
\vspace{-1.5em}
\end{table}

\paragraph{Is using larger context better?} \label{RQ2} To answer this question, we compare model performance on full tables vs. single and Oracle tables.

\textit{1. Full Table vs Single Table.}
Table \ref{tab:zero-shot} shows that using a random table as a prompt significantly doubles the performance of all models. This suggests that, although a single table does not provide any temporally changing information, the models might be accessing their pre-trained knowledge to answer questions accurately. Additionally, utilizing all the tables enhances the performance of all models (almost by 30-40\% in most cases), suggesting that the current LLMs can understand temporally evolving information. Providing a single table—specifically, the most recent table in the timeline—as context improves EM results for all models. In the few-shot setting, all models show improved exact match performance.

\textit{2. Full Tables v.s. Oracle Table.}
Oracle tables are retained from the dataset creation process and given to LLM as context for QA. The F1 score difference between the full timeline and oracle tables without decomposition is notable, with a gap of approximately 11.5$\%$ (GPT-4o) points in the zero-shot setting. This disparity suggests that LLMs struggle with temporal grounding, often failing to effectively extract relevant information from the timeline. However, the gap is significantly reduced to (4.0$\%$) by decomposing the tasks of information extraction and reasoning into multiple stages.

\paragraph{Task Decomposition Helps.} \label{RQ3}
When comparing results where Full Timeline as the context and task decomposition is used (Tables \ref{tab:zero-shot}, \ref{tab:few-shot}, and \ref{tab:cot}), we see that task decomposition further improves performance across all models by 10-20\%. Task decomposition prompting enhances the model's temporal grounding and improves its capacity to retrieve pertinent information for accurately answering questions. This improvement is consistently observed across various settings, including zero-shot, few-shot, and chain-of-thought approaches. We observe that \textit{Information Retrieval-Extraction} achieves the highest F1 and EM scores. However, \textit{Information Extraction}, which retrieves the correct table from a set of tables, closely follows in performance across most models, even outperforming others in specific instances, such as Llama3-8b and Gemini-1.5 Flash in both Zero-Shot and Chain-of-Thought scenarios. We observe a similar trend in Table \ref{tab:splits_results_tasks} across various reasoning types (as listed in Table \ref{tab:reasoning-split}), indicating that task decomposition consistently improves performance across all reasoning categories.

\textit{Iterative vs. Single Inference.} Furthermore, we observe task decomposition with multiple inference requests, i.e., a multi-prompt iterative pipeline improves model performance. We observe that sequential LLM requests for individual tasks outperform sending a single request combining all the tasks. Using GPT-4o, we implemented a three-stage process: information retrieval, extraction, and reasoning, with outputs from each stage serving as context for the subsequent task. This approach improved performance from 50 to 52 on both F1 and EM metrics on the full timeline with Chain of Throught prompting, suggesting enhanced performance when the model focuses on single tasks sequentially.

\vspace{-0.5em}
\paragraph{Retrieval Performance on Oracle Tables.} To assess the impact of the evidence retrieval approach, we compare retrieved tables with oracle tables. The results, presented in Table \ref{tab:retrieval_metrics} and Table \ref{tab:qa_pairs@k}, evaluate GPT-4o's performance using precision and recall metrics, comparing tables extracted from the Full Timeline against Oracle Tables. The IRE setting achieves a higher recall (95.25\%) than IR (63.15\%) for single-table retrieval while maintaining comparable precision. However, as the number of tables increases, GPT-4o struggles in both settings. This decline in performance with increasing table complexity highlights the need for more advanced techniques to improve multi-table retrieval. The performance drop in QA is expected since fewer tables are retrieved from the larger complete timeline as K decreases, making the QA task more challenging.

\paragraph{Reasoning Category-Wise Analysis.} \label{RQ5}
Table \ref{tab:splits_results_IE} presents the performance of different reasoning category splits (Table \ref{tab:reasoning-split}) when using Oracle Tables as context versus the Full Timeline for the information-extraction task decomposition. The results indicate that, on average, the Oracle context outperforms the Full Timeline. However, for specific reasoning categories such as \textit{Difference, Counting,} and \textit{Compare}, the full-timeline can achieve slightly better performance. See Tables \ref{tab:splits_results} in Appendix \ref{sec:appendix} for reasoning category-wise results on full vs. Oracle table without decomposition.

\begin{table}[!htb]
\vspace{-0.5em}
    \centering
    \scalebox{0.68}{
    \begin{tabular}{l|cc|cc}
    \textbf{Context} &\multicolumn{2}{c|}{\textbf{Full Timeline}} &\multicolumn{2}{c}{\textbf{Oracle Tables}} \\\cmidrule{1-5}
    
    &F1 &EM &F1 &EM \\\midrule
    \multicolumn{5}{c}{\textbf{Time Information}} \\\midrule
    Implicit               & 59.78                & 55                  & \textbf{62.99}                & \textbf{56.89}               \\
Explicit               & 62.71                & 55.22               &\textbf{65.36}                & \textbf{56.38}               \\\midrule
    \multicolumn{5}{c}{\textbf{Reasoning Types}} \\\midrule
    Extraction             & 67.87                & 62.62               & \textbf{72.51}                & \textbf{66.4}                \\
Percentage             & 65.15                & 56.15               & \textbf{67.15}                & \textbf{66.4}                \\
Difference             & \textbf{52.66}                & \textbf{46.37}               & 50.27                & 39.94               \\
Difference \& Compare  & \textbf{59.63}                & \textbf{56.51}               & 57.4                 & 52.58               \\
Counting               & \textbf{54.43}                & \textbf{54.29}               & 52.86                & 47.96               \\
Minimum                & 70.22                & 67.77               & \textbf{75.22}                & \textbf{72.72}               \\
Ratio                  & 56.2                 & 56.2                & \textbf{75.3}                 & \textbf{75.3}                \\
Maximum                & 57.4                 & \textbf{56.27}               & \textbf{58.37}                & 55.93               \\
Compare                & \textbf{69.26}                & \textbf{64.32}               & 64.58                & 61.52               \\\midrule
    \multicolumn{5}{c}{\textbf{\# of keys involved across timeline}} \\\midrule
    Multiple Keys          & 59.85                & 53.36               & \textbf{62.31}                & \textbf{54.76}               \\
Single Key             & 64.96                & 61.06               & \textbf{66.84}                & \textbf{62.06}               \\
\end{tabular}
    }
    \vspace{-0.5em}
    \caption{\small \textbf{Reasoning Category-wise Results with Information Extraction task decomposition.}  \label{tab:splits_results_IE} Results of Full Timeline vs. Oracle Tables context setting with COT prompting on GPT-4o and Information Extraction task decomposition.}
    \vspace{-1.5em}
\end{table}

\paragraph{COT > Few Shot > Zero Shot.} Chain-of-thought (COT) prompting consistently demonstrated superior results compared to other modeling scenarios in both few-shot and zero-shot settings. Notably, the performance achieved with COT using task decomposition surpasses that of task decomposition in zero-shot and few-shot models.

\noindent \paragraph{Open source v.s. Closed source.} \label{RQ1}
GPT-4o consistently outperformed other models on our dataset, demonstrating its robust reasoning capabilities. Gemini-1.5-flash and Llama3-70B models showed comparable performance across most settings (53 vs 47 vs 45 F1 score for Zero-Shot Information Retrieval). Although closed-source models (accessible via API) are updated frequently and typically exhibit significant performance advantages over open-source models, the minimal differences observed in this dataset suggest that the task presented by the proposed dataset has not been adequately addressed by existing datasets in the literature. Mixtral and Llama-8b exhibit the weakest performance among the models tested, likely due to their smaller size. This limited capacity may have affected their ability to handle complex prompts effectively. See Tables \ref{tab:zero-shot-appendix}, \ref{tab:few-shot-appendix}, and \ref{tab:cot-appendix} in Appendix \ref{sec:appendix} for R-1, R-L scores across all the settings.

\paragraph{Finetuning enhances LLMs Performance.} We fine-tuned GPT-4o-mini using 100 and 1,000 samples from the dataset, i.e., a small subset of data, reserving the remaining samples for the evaluation. Our results (Tables \ref{tab:zero-shot-fine-tuned}, \ref{tab:few-shot-fine-tuned}, and \ref{tab:cot-fine-tuned}) demonstrate a significant performance improvement, with all models achieving F1 scores exceeding 70.0. Furthermore, we found that fine-tuned models across various task decompositions performed equally well, suggesting that once the model has been fine-tuned, prompt-based granular task decomposition may not be necessary, as the model has already acquired the capability to address the queries effectively. Models finetuned on 1000 samples performed better than 100 samples, indicating that to effectively solve the problem of temporal reasoning, LLMs require a large amount of data. For more results of fine-tuned GPT-4o-mini checkout Tables \ref{tab:zero-shot-fine-tuned-rouge}, \ref{tab:few-shot-fine-tuned}, \ref{tab:cot-fine-tuned}, , \ref{tab:few-shot-fine-tuned-rouge}, and \ref{tab:cot-fine-tuned-rouge} in Appendix \ref{sec:appendix}.

\noindent

\paragraph{Temporal-Specific Models.} We evaluated recent temporal reasoning models against general-purpose language models, employing Chain of Thought (CoT) prompting with oracle tables and Key Extraction for task decomposition (see Table \ref{tab:temporal_model_comparison}). Our findings reveal that temporal-specific models (Timo-7B \cite{su2024timo}, Timellama-7B-Chat \cite{10.1145/3589334.3645376}) achieve notable improvements over comparable baseline models, demonstrating the effectiveness of temporal-focused post-training. However, they still trail larger general-purpose models, highlighting the dominance of scale over temporal specialization and the need for future research in integrating both advantages.
\begin{table}[!htb]
\centering
\scalebox{0.75}{
\begin{tabular}{l|rr}
\toprule
\textbf{Model} & \textbf{F1} & \textbf{EM} \\
\midrule
\multicolumn{3}{c}{\textbf{Larger General-Purpose Models}} \\ \midrule
GPT-4o & \textbf{60.44} & \textbf{56.56} \\
GPT-4o-Mini & 44.38 & 36.00 \\
Llama3-70B & 48.67 & 44.00 \\
Gemini-1.5-Flash & 48.23 & 42.89 \\
\midrule
\multicolumn{3}{c}{\textbf{Temporal-Specific Models}} \\ \midrule
Timo-7B & 26.16 & 23.02 \\
Timellama-7B-Chat & 25.99 & 21.78 \\
\midrule
\multicolumn{3}{c}{\textbf{Smaller Baseline Models}} \\ \midrule
Llama3-8B & 16.27 & 11.33 \\
Mixtral 8×7B & 15.28 & 11.89 \\
\bottomrule
\end{tabular}}
\vspace{-0.25em}
\caption{\textbf{Temporal Specific Models.} Performance comparison of models trained with temporal-focused post-tuning with comparable size models and other models.}
\label{tab:temporal_model_comparison}
 \vspace{-1.5em}
\end{table}

\section{Related Works}
\paragraph{Tabular Reasoning.}{Various NLP tasks on semi-structured tabular data have emerged as challenging due to the nature of the data \cite{gupta-etal-2020-infotabs}. Some of these include fact verification \cite{Chen2019TabFactAL, 10.1145/3357384.3357932, gupta-etal-2020-infotabs}, question answering, semantic parsing \cite{7845035, 10.1145/2872427.2883080, chen-etal-2020-hybridqa, lin-etal-2020-bridging, zayats-etal-2021-representations, oguz-etal-2022-unik, chen2021open, iyyer-etal-2017-search, krishnamurthy-etal-2017-neural, 10.1145/3397271.3401205, pasupat-liang-2015-compositional, 10.1145/3372117}, information synchronization \cite{khincha-etal-2023-infosync} and table-to-text generation \cite{parikh-etal-2020-totto, li-etal-2021-twt-table, nan-etal-2021-dart, yoran-etal-2022-turning, chen-etal-2020-logical}. A range of datasets and models have been developed to understand semi-structured information such as Table2vec \cite{10.1145/3331184.3331333}, TAPAS \cite{herzig-etal-2020-tapas}, TaBERT \cite{yin-etal-2020-tabert}, TabStruc \cite{zhang-etal-2020-table}, TABBIE \cite{iida-etal-2021-tabbie}, TabGCN \cite{pramanick-bhattacharya-2021-joint}, RCI \cite{glass-etal-2021-capturing} and model fine-tuning techniques such as \citealp{yu-etal-2018-spider, eisenschlos-etal-2020-understanding, neeraja-etal-2021-incorporating, shankarampeta-etal-2022-enhancing}.  Works such as \citealp{akhtar-etal-2023-exploring, srivastava-etal-2024-evaluating} studied the numerical reasoning capabilities of LLMs on tabular data, and \citealp{gupta-etal-2022-model, gupta-etal-2022-right} explore right evidence extraction for reasoning.}

\paragraph{Temporal Reasoning: }{Temporal question answering datasets such as TORQUE \cite{ning-etal-2020-torque}, TIMESENSITIVEQA \cite{chen2021a} focus on entity-specific reading comprehension with time-sensitive
questions created from Wikipedia paragraphs, SYGMA \cite{neelam-etal-2022-sygma}, CRONQUESTIONS \cite{saxena-etal-2021-question}, and TEMPQUESTIONS \cite{10.1145/3184558.3191536} explore question answering on temporal links in knowledge graph embeddings. Other temporal datasets such as SituatedQA\cite{zhang-choi-2021-situatedqa} explores open-domain question answering, TEMPLAMA \cite{dhingra-etal-2022-time} studies close-form questions. Moreover, work such as TempTabQA \cite{gupta-etal-2023-temptabqa}, TIQ \cite{10.1145/3589335.3651895}, TRAM \cite{wang2024tram}, BIG-bench \cite{srivastava2023beyond}  explores temporal reasoning on structured and semi-structured data. 

In contrast to previous studies such as \citealp{gupta-etal-2023-temptabqa} and \citealp{deng2024enhancing}, which primarily focus on single tables for a given entity, and those such as \citealp{10.1145/3589334.3645547} and \citealp{10.1145/3589335.3651895} that explore temporal question answering with implicit time constraints derived from diverse sources such as knowledge bases, text, and infoboxes, our research uniquely investigates the temporal reasoning capabilities of LLMs. Specifically, we examine how LLMs handle multiple tables related to a single entity over time, incorporating the evolving information within those tables. This involves extracting relevant evidence, comprehending the changing temporal context, and employing temporal reasoning skills to answer the questions posed.

\section{Conclusion and Future Work}
In conclusion, our study reveals key limitations in NLP systems' ability to reason about transient information in semi-structured data. We introduce a novel task of question answering on temporally evolving tables, along with a new \datasetName dataset containing 3,971 question-answer pairs from over 14k tables and 1,238 entities across various time periods. Evaluating state-of-the-art models on this dataset highlights shortcomings in evidence extraction and reasoning, underscoring the need for improved temporal reasoning in NLP models and guiding future research. \textbf{Future Direction.} (a) \textbf{Diverse Structures}: We plan to expand dynamic temporal QA beyond traditional tables to include hybrid formats with text, images, and graphs, as well as hierarchical structures capturing nested temporal data. This will better reflect real-world scenarios and improve model applicability. (b) \textbf{Neuro-symbolic Learning}: We aim to develop more robust, interpretable models by integrating neural networks with symbolic reasoning, enhancing accuracy and explainability in handling complex temporal queries.

\section*{Limitations}
This study's scope was confined to Wikipedia Infoboxes, limiting our findings' generalizability. Future research should encompass diverse table formats to provide a more comprehensive understanding.  Resource constraints restricted our fine-tuning process to a modest dataset of 1,000 samples. To gain a more nuanced understanding of the benefits of data-driven fine-tuning, it is crucial to examine the effects of this process on larger, more diverse datasets. It's important to note that the LLMs employed in this study were pre-trained on Wikipedia data, potentially introducing bias due to prior knowledge of the entities in our dataset. These limitations underscore the need for future work to address these constraints, enabling a more thorough evaluation of our proposed approach. 

A key limitation of our current evaluation is that we did not systematically assess the models' reliance on pre-trained knowledge by testing performance on data generated after their respective training cutoff dates. Such an analysis would require careful curation of a temporally stratified test set, identifying questions that reference post-cutoff information, and comparing performance across temporal splits. This type of evaluation could provide valuable insights into how models adapt to new information versus relying on pre-trained knowledge. While important, this analysis presents significant methodological challenges, including controlling for different cutoff dates across models and ensuring fair comparison conditions. We leave this systematic temporal evaluation as an important direction for future work.

Currently, our experiments are conducted in a closed-world setting, where entity-specific tables are directly associated with the query. This contrasts with an open-world retrieval setting, where relevant tables must be retrieved from a large corpus (e.g., Wikipedia) containing distractors. While closed-world evaluation simplifies table access, open-domain retrieval introduces more realistic challenges and remains an important direction for future work. Furthermore, our experiments were conducted solely using English-language data, allowing for expansion into multilingual contexts to assess the approach's efficacy across various languages. Subsequent studies should aim to overcome these boundaries, thereby enhancing the robustness and applicability of our findings across different domains and linguistic contexts.

\section*{Ethics Statement}
Our study examines how different language models (LMs) perform temporal reasoning with temporally evolving tabular data. We acknowledge that real-world applications of these systems require further testing specific to each use case. We uphold high ethical standards in our research and publication process. We provide complete details on datasets and evaluation methodologies to ensure our work can be reproduced. To support future work, we will share all scripts and resources used for creating the dataset and evaluating models. This promotes continued research in the field. We are dedicated to using computational linguistics methods responsibly and fairly. Our paper's claims accurately reflect our experimental results.  We used AI tools to assist us with writing, but we carefully checked and removed any errors or biases.

\section*{Acknowledgements}
Research was sponsored by the Army Research Office and was accomplished under Grant Number
W911NF-20-1-0080. The views and conclusions contained in this document are those of the authors and should not be interpreted as representing the official policies, either expressed or implied, of the Army Research Office or the U.S. Government. The U.S. Government is authorized to reproduce and distribute reprints for Government purposes notwithstanding any copyright notation herein. This work was partially funded by ONR Contract N00014-23-1-2364. We extend our gratitude to the annotators who verified our data and corresponding question answer pairs. We extend our sincere appreciation to Jennifer Sheffield from the University of Pennsylvania for her administrative support. Lastly, we extend our appreciation to the reviewing team for their insightful comments.

\bibliography{custom}
\clearpage
\appendix

\section{Appendix}
\label{sec:appendix}
\subsection{Human Evaluation} \label{subsec:human_eval_info}
We presented five evaluators with a unique set of 50 curated questions to establish a baseline for human performance. All evaluators were fluent English speakers and possessed graduate-level qualifications. In constructing the set of questions, we prioritized the selection of questions that required diverse reasoning operations in multiple tables in the timeline. This was aimed at ensuring a set of complex and high-quality questions for each evaluator. The evaluators were instructed to provide concise and direct answers without additional explanation. The final human performance baseline was determined by averaging the scores between all evaluators.

\subsection{Reasoning Category-Wise Analysis with no task decomposition}
Table \ref{tab:splits_results} indicates that the performance of various reasoning category splits (Table \ref{tab:reasoning-split}) in the Oracle Tables context setting is lower than in the Full Timeline context setting. This indicates that, despite providing the correct evidence (i.e., the exact tables necessary to address the question), GPT-4o still underperformed. Ideally, we would expect the Oracle context setting to yield the highest performance. This observation suggests that both extraction and reasoning processes may not be functioning correctly, potentially leading the model to rely on spurious correlations in its responses. A similar observation is also reported by \citealp{gupta-etal-2022-model}.

\begin{table}[!htb]
    \centering
    \scalebox{0.68}{
    \begin{tabular}{l|cc|cc}
    \textbf{Context} &\multicolumn{2}{c|}{\textbf{Full Timeline}} &\multicolumn{2}{c}{\textbf{Oracle Tables}} \\\cmidrule{1-5}
    
    &F1 &EM &F1 &EM \\\midrule
    \multicolumn{5}{c}{\textbf{Time Information}} \\\midrule
Implicit &\textbf{61.9}&\textbf{56} &58.23 &49.22 \\
Explicit &\textbf{65.94} &\textbf{55.88} &65.81 &\textbf{60.13} \\\midrule
\multicolumn{5}{c}{\textbf{Reasoning Types}} \\\midrule
Extraction &\textbf{69.84} &\textbf{63} &66.48 &57.62 \\
Percentage &\textbf{63.65} &\textbf{55.4} &49.12 &49.12 \\
Difference &\textbf{60.07} &\textbf{44.6} &41.27 &29.37 \\
Difference \& Compare &\textbf{58.54} &\textbf{57.2} &54.16 &49.58 \\
Counting &\textbf{58.88} &\textbf{58.08} &56.44 &55.62 \\
Minimum &\textbf{66.14} &\textbf{63.72} &54.28 &51.6 \\
Ratio &\textbf{74.3} &\textbf{74.3} &40.65 &40.65 \\
Maximum &\textbf{64.48} &\textbf{61.32} &57.6 &53.27 \\
Compare &\textbf{61.12} &58.12 &59.92 &\textbf{58.32} \\\midrule
\multicolumn{5}{c}{\textbf{\# of keys involved across timeline}} \\\midrule
Multiple Keys &\textbf{65.49} &\textbf{56.62} &57.93 &46.36 \\
Single Key &\textbf{67.15} &\textbf{61.56} &63.89 &58.26 \\
\end{tabular}
    }
    \vspace{-0.25em}
    \caption{\small \textbf{Reasoning Category-wise Results.}  \label{tab:splits_results} Results of Full Timeline vs Oracle Tables context setting with COT prompting on GPT-4o and without task decomposition.}
    \vspace{-1.0em}
\end{table}

\begin{table}[!htb]
    \centering
    \scalebox{0.65}{
    \begin{tabular}{l|cc|cc|cc}
\textbf{Decomposition} &\multicolumn{2}{c|}{\textbf{WD}} &\multicolumn{2}{c|}{\textbf{IE}} &\multicolumn{2}{c}{\textbf{IRE}} \\\cmidrule{1-7}
&F1 &EM &F1 &EM &F1 &EM \\\midrule
\multicolumn{7}{c}{\textbf{Reasoning Types}} \\\midrule
Extraction             & 69.84          & 63             & 67.87          & 62.62          & \textbf{76.86}           & \textbf{71.2}           \\
Percentage             & \textbf{63.65}          & 55.4           & 65.15          & 56.15          & 63.4            & \textbf{55.9}           \\
Difference             & 60.07          & 44.6           & 52.66          & 46.37          & \textbf{60.23}           & \textbf{52.09}          \\
Difference \& Compare  & 58.54          & 57.2           & 59.63          & 56.51          & \textbf{62.18}           & \textbf{60.45}          \\
Counting               & \textbf{58.88}          & \textbf{58.08}          & 54.43          & 54.29          & 57.53           & 57.58          \\
Minimum                & 66.14          & 63.72          & 70.22          & 67.77          & \textbf{74.06}           & \textbf{71.52}          \\
Ratio                  & 74.3           & 74.3           & 56.2           & 56.2           & \textbf{80.8}            & \textbf{78.8}           \\
Maximum                & 64.48          & 61.32          & 57.4           & 56.27          & \textbf{64.7}            & \textbf{63.52}          \\
Compare                & 61.12          & 58.12          & 69.26          & 64.32          & \textbf{65.18}           & \textbf{60.32}          \\\midrule
\multicolumn{7}{c}{\textbf{\# of keys involved across timeline}} \\\midrule
Multiple Keys          & 65.49          & 56.62          & 59.85          & 53.36          & \textbf{68}              & \textbf{60.96}          \\
Single Key             & 67.15          & 61.56          & 64.96          & 61.06          & \textbf{70.55}           & \textbf{66.56} \\
\end{tabular}
    }
    \caption{\small \textbf{Reasoning Category-wise Results across task decompositions.}  \label{tab:splits_results_tasks} Results of Full Timeline context setting with COT prompting on GPT-4o across various task decompositions.}
\end{table}

\subsection{Retrieval Metrics}
Table \ref{tab:retrieval_metrics} shows the retrieval performance metrics comparing Information Retrieval-Extraction (IRE) and Information Retrieval (IR) approaches across different numbers of retrieved tables (K). The metrics include F1 scores (for the question answering task), precision, and recall measured against Oracle tables. Table \ref{tab:qa_pairs@k} presents the distribution of question-answer pairs across different numbers of tables extracted for answering in Oracle, Information Retrieval (IR), and Information Retrieval-Extraction (IRE) settings.

\begin{table}[!htb]
     \centering
     \scalebox{0.57}{
     \begin{tabular}{p{2.3cm}@{}|c|ll|rr|rr}
& Samples used & \multicolumn{2}{c|}{\shortstack{\bf Without \\ \bf Fine Tuning}}&\multicolumn{2}{c|}{\bf 100} &\multicolumn{2}{c}{\bf 1000}     \\
\hline
\bf Context &\bf Decomposition  & R-1&R-L&R-1&R-L&R-1&R-L\\\midrule
No Table &-  & 21.98&21.86& 20.47&20.23&\textbf{23.75}&\textbf{23.75}\\\midrule
Single Table&-  & 31.09&31.09&38.38&38.43&\textbf{46}&\textbf{46.03}\\

\multirow{4}{*}{Full Timeline}&WD  & 43.8&42.92&54.04&53.16&\textbf{74.65}&\textbf{73.95}\\
&IE & 51.3&50.29&53.18&52.56& \textbf{75.86}& \textbf{75.18}\\
& IR & 48.53&47.43&53.1&52.54&\textbf{76.35}&\textbf{75.64}\\
 & IRE & 51.77&50.53& 52.76& 52.38& \textbf{75.3}& \textbf{74.79}\\
\end{tabular}}
\caption{\small\textbf{Zero-Shot Results with Fine Tuned GPT-4o-mini.} R-1 and R-L metrics in different in-context variations and intermediate task decompositions with zero-shot prompting.}
\label{tab:zero-shot-fine-tuned-rouge}
\end{table}

\begin{table}[!htb]
     \centering
     \scalebox{0.57}{
     \begin{tabular}{p{2.3cm}@{}|c|ll|rr|rr}
& Samples used & \multicolumn{2}{c|}{\shortstack{\bf Without \\ \bf Fine Tuning}}&\multicolumn{2}{c|}{\bf 100} &\multicolumn{2}{c}{\bf 1000}     \\
\hline
\bf Context &\bf Decomposition  & F1&EM&F1 &EM &F1 &EM \\\midrule
Single Table& -   & 35.31&30.92&38.95&34.5&\textbf{48.24}&\textbf{43.5}\\
Recent Table& -   & 35.31&36.08&54.86&54.9&\textbf{75.84}&\textbf{75.84}\\\midrule
\multirow{4}{*}{Full Timeline}&WD  & 43.73&37.01&52.28&47.3&\textbf{75.95}&\textbf{71.7}\\
&IE & 43.17&34.3&52.94&48& \textbf{76.79}& \textbf{72.9}\\
& IR & 45.24&38.7&52.18&47.2&\textbf{76.96}&\textbf{73.2}\\
 & IRE & 44.72&36.4& 52.27& 47.1& \textbf{76.85}& \textbf{73.2}\\
\end{tabular}}
\caption{\small \textbf{Few Shot Results with Fine Tuned GPT-4o-mini.} Results in different in-context variations and intermediate task decompositions with few-shot prompting.}
\label{tab:few-shot-fine-tuned}
\end{table}

\begin{table}[!htb]
     \centering
     \scalebox{0.7}{
     \begin{tabular}{c|ll|rr|rr}
Samples used & \multicolumn{2}{c|}{\shortstack{\bf Without \\ \bf Fine Tuning}}&\multicolumn{2}{c|}{\bf 100} &\multicolumn{2}{c}{\bf 1000}     \\
\hline
\bf Decomposition  & F1&EM&F1 &EM &F1 &EM \\\midrule
WD  & 49.06&41.79&52.77&47.9&\textbf{76.93}&\textbf{73.5}\\
IE  & 42.81&31.8&52.4&47.7& \textbf{76.95}& \textbf{73.1}\\
IR & 48.57&41.3&49.99&45&\textbf{77.27}&\textbf{73.3}\\
IRE & 56.4&48.33& 52.9& 47.56& \textbf{77.19}& \textbf{73.11}\\
\end{tabular}}
\caption{\small\textbf{COT Results with Fine Tuned GPT-4o-mini with Full timeline as context.} Results in different in-context variations and intermediate task decompositions with COT prompting.}
\label{tab:cot-fine-tuned}
\end{table}

\begin{table*}[!htb]
     \centering
     \scalebox{0.68}{
     \begin{tabular}{p{2.3cm}@{}|c|rr|rr|rr|rr|rr|rr}
& &\multicolumn{2}{c|}{\bf GPT-4o} &\multicolumn{2}{c|}{\bf Llama3-70b} &\multicolumn{2}{c|}{\bf Gemini-1.5} &\multicolumn{2}{c|}{\bf GPT-4o-mini} &\multicolumn{2}{c|}{\bf Llama3-8b} &\multicolumn{2}{c
}{\bf Mixtral} \\\cmidrule{3-14}
\bf Context &\bf Decomposition &R-1&R-L&R-1&R-L&R-1&R-L&R-1&R-L&R-1&R-L&R-1&R-L\\\midrule

No Table & - & 21.98&21.86&13.42&13.3&15.69&15.52&21.98&21.86&11.2&11.09&9.53&9.43\\\midrule

Single Table& - &33.06&33.11&15.82&15.78&26.99&26.98&31.09&31.09&14.58&14.5&14.22&14.2\\\midrule

\multirow{4}{*}{Full Timeline}&Without Decompostion &48.8&47.91&47&45.89&39.56&38.47&43.8&42.92&39.04&38.47&31.94&31.19\\

&Information Retrieval&55.8&54.4& 45.71& 44.75&40.49&39.47&48.53&47.43& 22.24& 21.61& \bf 32.24& \bf 31.58\\

& Information Extraction&\bf 57.42&\bf 56.25&\bf 49.66&\bf 48.31&\bf 52.77&\bf 51.8&51.3&50.29&\bf 37.7&\bf 37.05&28.93&28.23\\

& Information Retrieval-Extraction& 56.81& 55.43& 49& 47.9& 50.4& 49.12& \bf 51.77& \bf 50.53& 27.72& 27.06& 29.1&28.3\\\midrule

\multirow{2}{*}{Oracle Tables}&Without Decompostion &61.17&60.59& 46.92& 46.34& 45.61& 45.02& 48.27&47.82& 40.16& 39.67& 25.72& 25.49\\

&Information Extraction &59.56&58.76& 42.56& 42.31& 43.7& 43.69&50.04&49.48& 14.77& 14.68& 26.4& 26.18\\
\end{tabular}}
\caption{\small\textbf{Zero Shot Results.} Results in different in-context variations and different intermediate task decompositions with zero-shot prompting. R-1 and R-L are reported for all models.}
\label{tab:zero-shot-appendix}
\end{table*}

\begin{table*}[!htb]
     \centering
     \scalebox{0.68}{
     \begin{tabular}{p{2.3cm}@{}|c|rr|rr|rr|rr|rr|rr}
& &\multicolumn{2}{c|}{\bf GPT-4o} &\multicolumn{2}{c|}{\bf Llama3-70b} &\multicolumn{2}{c|}{\bf Gemini-1.5} &\multicolumn{2}{c}{\bf GPT-4o-mini} &\multicolumn{2}{c|}{\bf Llama3-8b} &\multicolumn{2}{c
}{\bf Mixtral} \\\cmidrule{3-14}
\bf Context &\bf Decomposition &R-1&R-L&R-1&R-L&R-1&R-L&R-1&R-L&R-1&R-L&R-1&R-L\\\midrule
Single Table& -  &35.24&35.23&17.96&17.92&32.29&32.31&33.39&33.4&15.28&15.24&20.52&20.48\\\midrule

\multirow{4}{*}{Full Timeline}&Without Decompostion &57.9&56.69&48.7&47.6&50.52&49.42&47.32&46.34&30.97&30.48&\bf 34.75&\bf 33.93\\

&Information Retrieval&54.83&53.76& 49.57& 48.85&50.47&49.58&48.07&47.46& 25.58& 25.07&  33.99& 33.47\\

& Information Extraction&57.55&56.52&50.56&49.84&\bf 51.61&\bf 50.43&48.76&47.85&\bf 36.56&\bf 35.84&27.65&27.36\\

& Information Retrieval-Extraction& \bf 59.2& \bf 58.09& \bf 50.88& \bf 50.05& 51.22& 50.07& \bf 49.61& \bf 48.7& 23.39& 22.78& 26.05&25.64\\\midrule

\multirow{2}{*}{Oracle Tables}&Without Decompostion &65.97&65.31& 54.47& 53.89& 51.81& 51.2&53.16&52.51& 39.02& 38.74& 32.57& 32.06\\

&Information Extraction &60.42&59.62& 48.09& 47.64& 50.17& 49.58&49.56&49.03& 17.91& 17.68& 30.41& 29.86\\

\end{tabular}}
\caption{\small\textbf{Few Shot Results.} Results in different in-context variations and intermediate task decompositions with few-shot prompting. R-1 and R-L scores are reported for all models. }
\label{tab:few-shot-appendix}
\end{table*}

\begin{table*}[!htb]
     \centering
     \scalebox{0.68}{
     \begin{tabular}{p{2.3cm}@{}|c|rr|rr|rr|rr|rr|rr}
& &\multicolumn{2}{c|}{\bf GPT-4o} &\multicolumn{2}{c|}{\bf Llama3-70b} &\multicolumn{2}{c|}{\bf Gemini-1.5} &\multicolumn{2}{c}{\bf GPT-4o-mini} &\multicolumn{2}{c|}{\bf Llama3-8b} &\multicolumn{2}{c
}{\bf Mixtral} \\\cmidrule{3-14}
\bf Context &\bf Decomposition &R-1&R-L&R-1&R-L&R-1&R-L&R-1&R-L&R-1&R-L&R-1&R-L\\\midrule

\multirow{4}{*}{Full Timeline}&Without Decompostion &62.85&61.51&57.37&56.02&59.03&57.92&54.43&53.43&43.25&42.53&\bf 37.2&\bf 36.28\\

&Information Retrieval&63.67&62.56& 53.45& 52.66&53.77&52.97&52.06&51.37& 27.51& 26.8& 33.8& 33.19\\

& Information Extraction&65.64&64.42&\bf 59.51&\bf 58.38&58.51&57.68&52.79&51.9&\bf 38.23&\bf 37.62&29.05&28.65\\

& Information Retrieval-Extraction& \bf 69.51& \bf 68.14& 57.23& 55.99& \bf 63.81& \bf 62.88& \bf 60.98& \bf 59.94& 25.34& 24.68& 30.66&30.22\\\midrule

\multirow{2}{*}{Oracle Tables}&Without Decompostion &67.21&66.29& 58.42& 57.79& 60.22& 59.63&60.47&59.79& 38.23& 37.97& 27.64& 27.34\\

&Information Extraction &65.34&64.48& 52.13& 51.72& 53.34& 52.39&52.37&51.83& 18.29& 18.19& 16.89& 16.44\\

\end{tabular}}
\caption{\small\textbf{COT Results.}  Results in different in-context variations and intermediate task decompositions with chain-of-thought prompting. R-1 and R-L scores are reported for all models.}
\label{tab:cot-appendix}
\end{table*}

\begin{table}[!htb]
     \centering
     \scalebox{0.6}{
     \begin{tabular}{p{2.3cm}@{}|c|ll|rr|rr}
& Samples used & \multicolumn{2}{c|}{\shortstack{\bf Without \\ \bf Fine Tuning}}&\multicolumn{2}{c|}{\bf 100} &\multicolumn{2}{c}{\bf 1000}     \\
\hline
\bf Context &\bf Decomposition  & R-1&R-L&R-1&R-L&R-1&R-L\\\midrule

Single Table& -  & 33.39&33.4&40.2&40.21&\textbf{49.55}&\textbf{49.54}\\\midrule
\multirow{4}{*}{Full Timeline}&WD  & 47.32&46.34&54.17&53.41&\textbf{77.65}&\textbf{76.95}\\

&IE & 48.76&47.85&54.76&54.04&\textbf{78.36}&\textbf{77.58}\\

& IR & 48.07&47.46&54.25&53.48&\textbf{78.42}&\textbf{77.84}\\

& IRE & 49.61&48.7&54.05&53.49&\textbf{78.65}&\textbf{77.84}\\
\end{tabular}}
\caption{\small\textbf{Few Shot Results with Fine Tuned GPT-4o-mini.} R-1  and R-L metrics in different in-context variations and intermediate task decompositions with few-shot prompting.}
\label{tab:few-shot-fine-tuned-rouge}
\end{table}

\begin{table}[!htb]
     \centering
     \scalebox{0.6}{
     \begin{tabular}{p{2.3cm}@{}|c|ll|rr|rr}
& Samples used & \multicolumn{2}{c|}{\shortstack{\bf Without \\ \bf Fine Tuning}}&\multicolumn{2}{c|}{\bf 100} &\multicolumn{2}{c}{\bf 1000}     \\
\hline
\bf Context &\bf Decomposition  & R-1&R-L&R-1&R-L&R-1&R-L\\\midrule
\multirow{4}{*}{All Tables}&WD  & 54.43&53.43&54.73&54.17&\textbf{78.61}&\textbf{78.02}\\
&IE  & 52.79&51.9&54.43&53.83&\textbf{78.44}&\textbf{77.84}\\
&IR & 52.06&51.37&53.1&52.62&\textbf{78.76}&\textbf{77.96}\\
&IRE & 60.98&59.94&54.97&54.47&\textbf{78.94}&\textbf{78.19}\\
\end{tabular}}
\caption{\small\textbf{COT Results with Fine Tuned GPT-4o-mini.} R-1 and R-L metrics in different in-context variations and intermediate task decompositions with COT prompting.}
\label{tab:cot-fine-tuned-rouge}
\end{table}

\begin{table}[!htb]
\centering
\scalebox{0.8}{
\begin{tabular}{lccc}
\textbf{\# Of Tables} &\textbf{Oracle} &\textbf{IR} &\textbf{IRE} \\\midrule
1 &2186 &1518 &1288 \\
2 &930 &1102 &1196 \\
3 - 5 &136 &434 &644 \\
6 - 10 &379 &347 &533 \\
> 10 &70 &43 &40 \\
\bottomrule
\end{tabular}
}
\caption{\small Number of QA pairs retrieved \# number of tables from Full Timeline in Information Retrieval \& Information Retrieval-Extraction setting with COT prompting using GPT-4o.}\label{tab:qa_pairs@k}
\end{table}


\subsection{Model Hyper parameters}
The following are the hyperparameters used for various models in our experiments:
\begin{itemize}
    \item  GPT-4o \& GPT-4o-mini with default parameters: temperature: 1.0, top\_p: 1.0, presence\_penalty: 0.
    \item  Gemini-1.5-Flash with parameters: temperature: 1.0, top\_p: 0.95, top\_k: 64
    \item  Llama3-70b \& Llama3-8b with default parameters: temperature: 1.0, top\_p: 1.0, presence\_penalty: 0.
    \item  Mixtral with default parameters: temperature: 1.0, top\_p: 1.0, presence\_penalty: 0.
    
\end{itemize}

\subsection{Entity Category Results}
Results in table \ref{tab:category_wise_results} show that performance across all entity categories with task decomposition (information-extraction-retrieval). 

\begin{table}[!htb]
    \centering
        \scalebox{0.68}{
    \begin{tabular}{l|cc|cc}

&\multicolumn{2}{c|}{\textbf{w/o Decomposition }} &\multicolumn{2}{c}{\textbf{IRE decomposition}} \\      	
Category  &	F-1  &	EM  &	F-1 &	EM \\
\hline
Cyclist 	& 52.6 &	52.0 &	53.4 &	53.0 \\
Equestrian 	& 58.4 &	58.0 &	62.2 &	62.0 \\
Field hockey &	34.1 &	34.0 &	34.5 &	34.0  \\
Golfer 	& 37.3 	&37.0 &	56.7 &	57.0 \\
Table tennis player  &	52.8 &	53.0 &	52.3 &	52.0 \\
Country 	& 70.7 &	59.6 &	77.5 &	68.0 \\ 
Cricket team &	67.7 &	55.7 &	70.0 &	60.0 \\
Gov Agencies &	72.3 &	51.1 &	77.4 &	59.0 \\
Economy 	&52.0 &	42.0 &	54.3 &	44.0 \\
Cricketer 	&79.8  &	76.8 &	80.0 	&77.0 \\
\hline
Average 	&56.79& 	51.45 &	\bf 60.91 & 	\bf 56.14 
    \end{tabular}
    }
    \caption{\textbf{Performance Across Various Categories with full timeline setting for various categories.}}
    \label{tab:category_wise_results}
\end{table}

\subsection{QA Templates}
\label{appendix:qa_template}

These predefined templates are used to generate questions for the cricket team category. Similar question-generation templates for all the categories are already available in our dataset.\\

\textbf{Template 1} - Name the person(s) who served as the <coach/test coach/od coach/batting coach/bowling coach/fielding coach> when <captain/test captain/od captain/t20i captain:value1> was the <captain/test captain/od captain/t20i captain:key1>? \\

\textbf{Template 2} - Who was the <coach/test coach/od coach> when <captain/t20i captain/od captain/test captain:key1> was <captain/t20i captain/od captain/test captain: value1> and <batting coach/bowling coach/fielding coach:key2> was <batting coach/bowling coach/fielding coach: value2>? \\

\textbf{Template 3} - Does the Indian Cricket Team have the best win percentage in the <test/odi/t20i> format in <year:value1> or <year:value2>? \\

\textbf{Template 4} - In which year did the <captain/test captain/od captain/t20i captain: value> became the <captain/test captain/od captain/t20i captain> of the Indian Cricket Team for the first time?\\

\textbf{Template 5} - Which person had the <longest/shortest> tenure as the <test captain/od captain/t20i captain/captain :key> of the Indian Cricket Team? \\

\textbf{Template 6} - Who was the <test captain/od captain/t20i captain/captain> of the Indian Cricket Team <before/after> <test captain/od captain/t20i captain/captain: value>?\\

\textbf{Template 7} - How many <total matches(including ODIs,Tests,T20Is)/test/odi/t20> matches the Indian Cricket Team played between <year:value1> and <year:value2>?\\

\textbf{Template 8} - what was the best <test/odi/t20i> rank of the Indian Cricket Team in <year:value>?\\

\textbf{Template 9} - Name the people who served as <test captain/od captain/t20i captain/captain> of Indian Cricket Team between <year:value1> and <year:value2>?\\

\textbf{Template 10} - Based on the given timeline, how many people served as the <test captain/od captain/t20i captain/captain> of the Indian Cricket Team? \\

\begin{table*}[!htb]
\centering
\scalebox{0.49}{
\begin{tabular}{c|ccc|ccc|ccc|ccc|ccc|ccc}
K = \# of Tables &\multicolumn{3}{c|}{K=1} &\multicolumn{3}{c|}{K<=2} &\multicolumn{3}{c|}{K<=3} &\multicolumn{3}{c|}{K<=5} &\multicolumn{3}{c|}{K<=10} &\multicolumn{3}{c}{K>10} \\\cmidrule{2-19}
&QA &Precision &Recall &QA  &Precision &Recall &QA  &Precision &Recall &QA  &Precision &Recall &QA  &Precision &Recall &QA  &Precision &Recall \\\midrule
IRE &43.84 &90.44 &95.25 &71.29 &76.3 &79.99 &70.84 &76.45 &79.64 &70.42 &76.81 &79.58 &71.85 &78.2 &79.74 &90.11 &80.13 &50.9 \\
IR &42.94 &80.72 &63.15 &71.05 &76.1 &79.1 &71.05 &76.49 &78.95 &71.06 &76.9 &78.92 &72.46 &78.57 &78.91 &86.61 &91.1 &62.55 \\
\bottomrule
\end{tabular}
}
\caption{\small\textbf{Retrieval Metrics of GPT-4o in Information Retrieval-Extraction (IRE) \& Information Retrieval (IR) setting with COT prompting}. Precision and Recall are measured between the tables extracted from Full Timeline vs Oracle Tables. The QA is the F1 score of the final question-answering task after the table retrieval. }\label{tab:retrieval_metrics}
\end{table*}

\begin{table*}[!htb]
\small
\centering
\scalebox{0.9}{
\begin{tabular}{l|c|c|c|c|c}

\textbf{Dataset} & \textbf{QA pairs} & \textbf{Evidence Formats} & \textbf{Source} & \textbf{Annotation Method} & \textbf{Type of questions} \\
\midrule
TabQA & 11,454 & Single table & Wikipedia & Human & Implicit \& Explicit \\
TIQ & 10,000 & Single table/text/KB & Wikipedia & Automated & Implicit \\
TransientTables & 3,971 & Multi-table & Wikipedia & Automated & Implicit \& Explicit \\

\end{tabular}}
\caption{Comparison of Question-Answering Datasets}
\label{tab:qa-datasets}
\end{table*}

\subsection{Prompts Used for Experimentation}
\label{appendix:prompts}
In our implementation, we converted all tabular data to JSON string format before passing them to the LLM. Below are examples of the entire prompt input of the LLM in the CoT setting. Depending on the category the few-shot examples (content in \{\}) change accordingly.

\begin{tcolorbox}[title=Prompt for COT Information Retrieval-Extraction, breakable]
Perform the following tasks –\\

Task 1: For the question provided with the timeline, retrieve the relevant tables from the timeline that shall be used to answer the question. The task is to extract the appropriate tables rather than generate the answer to the question.\\

Task 2: From the tables retrieved in task 1, retrieve the relevant keys and values that would be used to answer the question.\\

Task 3: Answer the question using the retrieved keys from Task 2. The answer should be concise, within 5 to 10 words. Further, answer the question based solely on the information presented in the retrieved key(s) without referencing any external data or information.\\

Here’s an example for your reference –\\
Timeline: \{example\_timeline\}\\
question 1: {question1}\\
Task 1 Answer: \{task1\_answer1\}\\
Task 2 Answer: \{task2\_answer1\}\\
Task 3 Answer: \{task3\_answer1\}\\
question 2: \{question2\}\\
Task 1 Answer: \{task1\_answer2\}\\
Task 2 Answer: \{task2\_answer2\}\\
Task 3 Answer: \{task3\_answer2\}\\
question 3: \{question3\}\\
Task 1 Answer: \{task1\_answer3\}\\
Task 2 Answer: \{task2\_answer3\}\\
Task 3 Answer: \{task3\_answer3\}\\

Now, perform the tasks for the following timeline(premise) and question -  \\
Premise: \{timeline\}\\
Question: \{question\}\\
Provide answers for task 1, task 2, and task 3 separately. Also, give a final answer based on the reasoning in task 3. For task 1, just retrieve the timestamps of the relevant tables.\\
Task 1 Answer:\\
Task 2 Answer:\\
Task 3 Answer:\\
Final Answer:\\\\
\end{tcolorbox}
\begin{tcolorbox}[title=Prompt for COT Information Extraction, breakable]
Perform the following tasks \\

Task 1: For the question provided with the timeline, retrieve the relevant keys from the relevant tables in the timeline that shall be used to answer the question. The task is to extract the appropriate keys from the relevant tables rather than generate the answer to the question.\\

Task 2: Answer the question using the retrieved keys from Task 1. The answer should be concise, within 5 to 10 words. Further, answer the question based solely on the information presented in the retrieved key(s) without referencing any external data or information.\\

Here’s an example for your reference –\\
Timeline: \{example\_timeline\}\\
question 1: \{question1\}\\
Task 1 Answer: \{task1\_answer1\}\\
Task 2 Answer: \{task2\_answer1\}\\
question 2: \{question2\}\\
Task 1 Answer: \{task1\_answer2\}\\
Task 2 Answer: \{task2\_answer2\}\\
question 3: \{question3\}\\
Task 1 Answer: \{task1\_answer3\}\\
Task 2 Answer: \{task2\_answer3\}\\

Now, perform the tasks for the following timeline(premise) and question -  \\
Premise: {timeline}\\
Question: {question}\\
Provide answers for task 1,and task 2 separately. Also, give a final answer based on the reasoning in task 2. \\
Task 1 Answer:\\
Task 2 Answer:\\
Final Answer:\\\\
\end{tcolorbox}

\begin{tcolorbox}[title=Prompt for COT Information Retrieval, breakable]
Perform the following tasks –\\
Task 1: For the question provided with the timeline, retrieve the relevant tables from the timeline that shall be used to answer the question. The task is to extract the appropriate tables rather than generate the answer to the question.\\

Task 2: Answer the question using the retrieved tables from Task 1. The answer should be concise, within 5 to 10 words. Further, answer the question based solely on the information presented in the retrieved table(s) without referencing any external data or information.\\

Here’s an example for your reference –\\
Timeline: \{example\_timeline\}\\
question 1: \{question1\}\\
Task 1 Answer: \{task1\_answer1\}\\
Task 2 Answer: \{task2\_answer1\}\\
question 2: \{question2\}\\
Task 1 Answer: \{task1\_answer2\}\\
Task 2 Answer: \{task2\_answer2\}\\
question 3: \{question3\}\\
Task 1 Answer: \{task1\_answer3\}\\
Task 2 Answer: \{task2\_answer3\}\\

Now, perform the tasks for the following timeline(premise) and question -  \\
Premise: \{timeline\}\\
Question: \{question\}\\
Provide answers for task 1, and task 2 separately. Also, give a final answer based on the reasoning in task 2. \\
Task 1 Answer:\\
Task 2 Answer:\\
Final Answer:\\
\end{tcolorbox}
\newpage

\end{document}